\documentclass[5p]{elsarticle} 

\usepackage{lineno,hyperref}
\modulolinenumbers[5]

\journal{Journal of \LaTeX\ Templates}

\usepackage{graphicx,color}
\usepackage{amsmath,amssymb}

\newcommand{\T}{^{^{_T}}}
\usepackage{enumitem}
\usepackage{algorithm} 
\usepackage{algorithmic}
\usepackage{subcaption}

\usepackage{array}
\newcommand{\PreserveBackslash}[1]{\let\temp=\\#1\let\\=\temp}
\newcolumntype{C}[1]{>{\PreserveBackslash\centering}p{#1}}
\newcolumntype{R}[1]{>{\PreserveBackslash\raggedleft}p{#1}}
\newcolumntype{L}[1]{>{\PreserveBackslash\raggedright}p{#1}}

\begin{document}

\begin{frontmatter}

\title{Extreme Learning Machine design for dealing with unrepresentative features}


\author[sinc,imal]{Nicol\'as Nieto}\ead{nnieto@sinc.unl.edu.ar}
\author[sinc]{Francisco Ibarrola}
\author[imal]{Victoria Peterson}
\author[sinc]{Hugo L. Rufiner}
\author[imal]{Ruben Spies}

\address[sinc]{Instituto de Investigaci\'on en Se\~nales, Sistemas e Inteligencia Computacional, sinc(i), UNL-CONICET, FICH, Ciudad Universitaria, CC 217, Ruta Nac. 168, km 472.4, (3000) Santa Fe, Argentina.}
\address[imal]{Instituto de Matem\'{a}tica Aplicada del Litoral, IMAL,  UNL-CONICET, Centro Cient\'ifico Tecnol\'ogico CONICET Santa Fe, Colectora Ruta Nac. 168, km 472, Paraje ``El Pozo'', (3000), Santa Fe, Argentina.}


\begin{abstract}
Extreme Learning Machines (ELMs) have become a popular tool in the field of Artificial Intelligence due to their very high training speed and generalization capabilities. Another advantage is that they have a single hyper-parameter that must be tuned up: the number of hidden nodes. Most traditional approaches dictate that this parameter should be chosen smaller than the number of available training samples in order to avoid over-fitting. In fact, it has been proved that choosing the number of hidden nodes equal to the number of training samples yields a perfect training classification with probability 1 (w.r.t. the random parameter initialization). In this article we argue that in spite of this, in some cases it may be beneficial to choose a much larger number of hidden nodes, depending on certain properties of the data. We explain why this happens and show some examples to illustrate how the model behaves. In addition, we present a pruning algorithm to cope with the additional computational burden associated to the enlarged ELM. Experimental results using electroencephalography (EEG) signals show an improvement in performance with respect to traditional ELM approaches, while diminishing the extra computing time associated to the use of large architectures.
\end{abstract}


\end{frontmatter}


\section{Introduction}\label{intro}

The use of random weights in Neural Networks was first proposed by Schmidt \emph{et al} in \cite{schmidt1992feed}. The idea was then reintroduced under the concept of Extreme Learning Machines (ELMs) in 2006 by Huang \emph{et al} (\cite{huang2004extreme}), and has been widely used by the Machine Learning community ever since \cite{huang2011extreme,wong2014real,zhang2018multi,song2013automatic}. This is so mainly because the training process is very fast and they have good generalization capabilities.

Another quite appealing aspect of ELMs is that, unlike most of the machine learning techniques, they have only one hyper-parameter that must be tuned up: the number of hidden nodes. Furthermore, Huang and Babri \cite{huang1998upper} have proved that given a training dataset consisting of $N$ samples, the model can learn them exactly using $N$ hidden nodes.
Nevertheless, perfect classification over a training set usually entails a loss of generalization capabilities. In fact, it is widely accepted that given $N$ training samples, $N$ is an upper bound for the number of nodes that can be chosen for the hidden layer (\cite{huang1998upper}).

In this work we argue that while choosing the number of hidden nodes $M$ close to $N$ is indeed a bad idea, choosing $M>>N$ may result in a significantly better performance. In the next section, we provide an explanation to why such a choice might improve performance, and run an experiment to attest it. A post-training pruning method in order to mitigate the extra computational burden in the training stage, coming from the choice of a larger number of hidden nodes, is described in Section \ref{sec:pruning}.

\section{Extreme Learning Machines}\label{ELM}

For simplicity, we shall consider an ELM within the context of a binary classification problem. We point out, however, that all the results presented in this work remain valid for multi-class problems.

Given an arbitrary vector $x \in \mathbb{R}^D$, the ELM classification output is given by
\begin{align}
z = \beta^T g(W x+b),    
\end{align}
where $W \in \mathbb{R}^{M\times D}$ is the matrix associated to the hidden layer, $g:\mathbb{R}\rightarrow \mathbb{R}$ is an activation function, $b\in\mathbb{R}^M$ is the bias vector, and $\beta\in\mathbb{R}^M$ is the weight vector connecting the hidden layer to the output. Here and in the sequel, the action of $g$ on a vector or a matrix is meant to be its components-wise evaluation.

The training process of an ELM consists of two main steps. First, the entries of $W$ and $b$ are randomly chosen as independent realizations of an absolutely continuous random variable (usually with uniform distribution in $[-1, 1]$). The second step consists of finding an appropriate vector $\beta$. This can be done as described below.

Let us consider a dataset consisting of $N$ training samples $x_n \in \mathbb{R}^D$, $n = 1, \ldots, N$, stacked in a matrix $X \in \mathbb{R}^{D \times N}$. Let $y\in\{-1,1\}^N$ be the desired output vector, where $y_n$, gives account for the class of $x_n, \;\forall n$.

Let us define
\begin{align}
H \doteq [g(W X+b\,1_{(1,N)})]\T,    
\end{align}
where $1_{(1,N)}$ is a row vector with all its elements equal to 1. Then, training the ELM weight vector $\beta$ amounts to solving the linear system
\begin{align}\label{eqn:Hb=y}
H \beta = y.    
\end{align}
Note that $H\in \mathbb{R}^{N \times M}$, meaning that the linear system is under-determined if $N>M$, e.g. the number of nodes is less than the number of training samples. In this case, we can define the vector $\hat{\beta}$ as a minimal-norm least-squares approximate solution of (\ref{eqn:Hb=y}). That is
\begin{align}\label{eqn:b_hat}
\hat{\beta} \doteq H^{\dagger}y,
\end{align}
where $H^{\dagger}$ is the Moore-Penrose generalized inverse.

When $M<N$, the ELM's generalization capability is associated with the fact that least squares solutions assign larger weights to the columns of $H$ which are most relevant for classification purposes. However, when $M=N$, system (\ref{eqn:Hb=y}) has a unique solution (with probability 1, as shown in  \cite{huang1998upper}). Hence, in this case, the solution $\hat{\beta}$ is forced to take all columns of $H$ into account, even those witch are irrelevant for classification proposes. This is a classic case of overfitting.

Although it is theoretically true that $M\geq N$ implies that the matrix $H$ has (with probability 1) $N$ independent column vectors, this does not necessary mean that the feature space be well represented. To illustrate this, let us consider the following simple example. Suppose that there are two columns of $H$, $h$ and $h' \in \mathbb{R}^N$, such that $h_1 = h'_1 +\epsilon$ (for a small $\epsilon\in \mathbb{R}$) and $h_n = h'_n, \forall n\geq 2$. Although these two vectors are strictly different, for all practical purposes they clearly encode the same feature information. More formally, while $M = N$ ensures that $H$ be invertible, it can still present very small singular values, which is a reflection of a poor representation of the feature space. As shown in \cite{horn1991topics}, adding random columns to a matrix increases its smaller non-zero singular value, which in a context of normalized data indicates a good representation of the feature space. 

\section{Changing the network size} \label{sec:netsize}

In light of the above discussion, we argue that, in order to optimize the network performance, $M$ should be chosen differently. Our proposal is training the ELM by choosing $M>>N$ followed by an \emph{a-posteriori} pruning method for reducing the size of the model. Although working with a large number of hidden nodes may increase the computing time, this method gets rid of the need for training the network several times for validating $M$. Hence, by an appropriate choice of the model, the final training time could end up being reduced, as it is shown in the next section.

\begin{figure}
\includegraphics[width=\columnwidth]{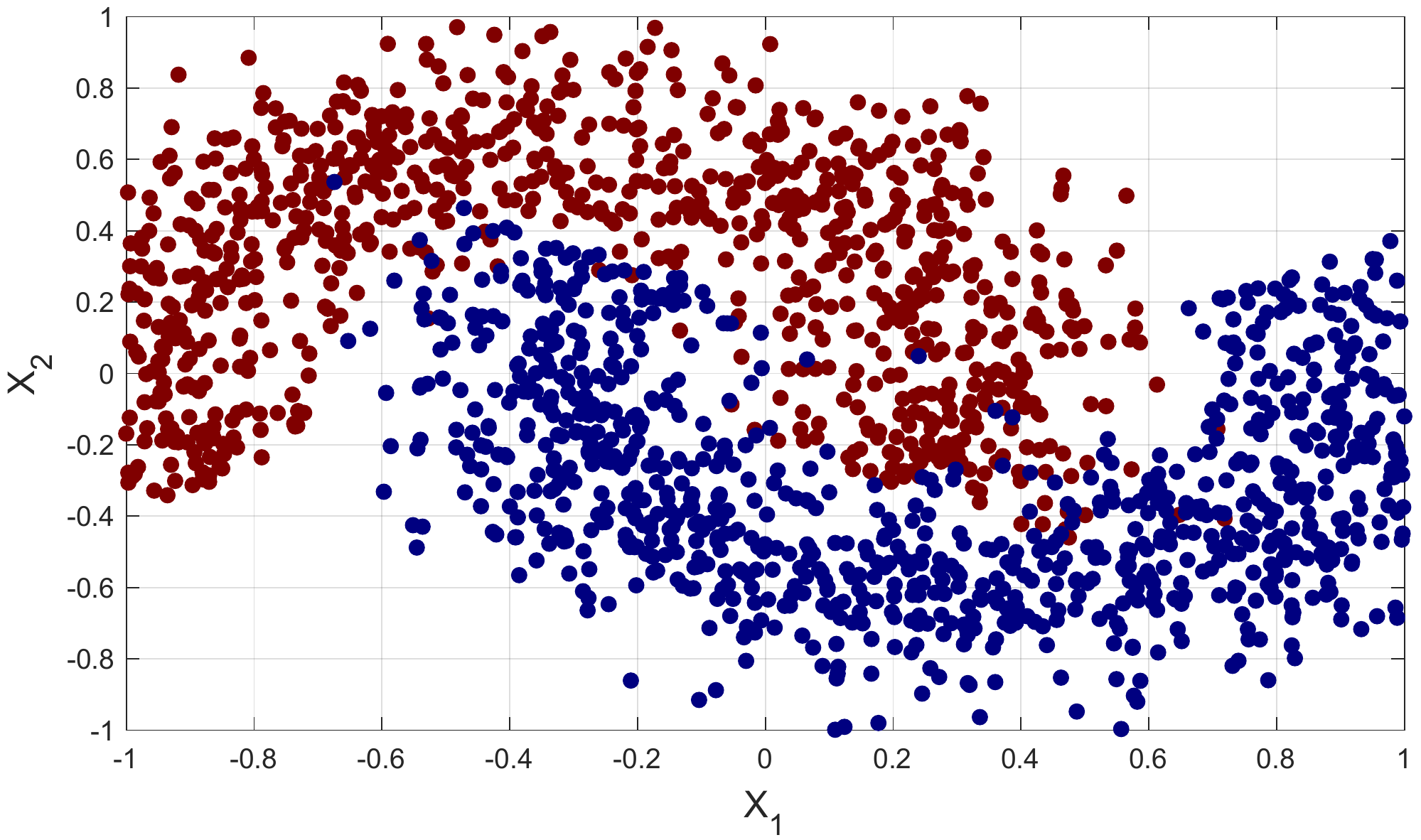}
\vspace{-0.5cm}
\caption{Synthetic class distribution}
\label{fig:lunitas}   
\vspace{-0.2cm}
\end{figure}

We now present a simple experiment using artificially generated data in order to illustrate the potential advantage of choosing $M>>N$.
Let us consider a binary classification problem in which the data points corresponding to the two classes are distributed as depicted in Figure \ref{fig:lunitas}. 
It is clear that the two displayed coordinates $x_1, x_2$ are enough for classification.
In an ideal situation, any additional coordinate added to the data points, taking random values independently of the class, should not be taken into account by the classifier. In a real problem, the data points might be highly contaminated with this kind of uninformative junk features (coordinates) which are, \emph{a-priori}, indistinguishable from the representative ones. Hence a question that immediately arises is: should the same kind of ELM architecture be used for this kind of problems?

\begin{figure}
\includegraphics[width=\columnwidth]{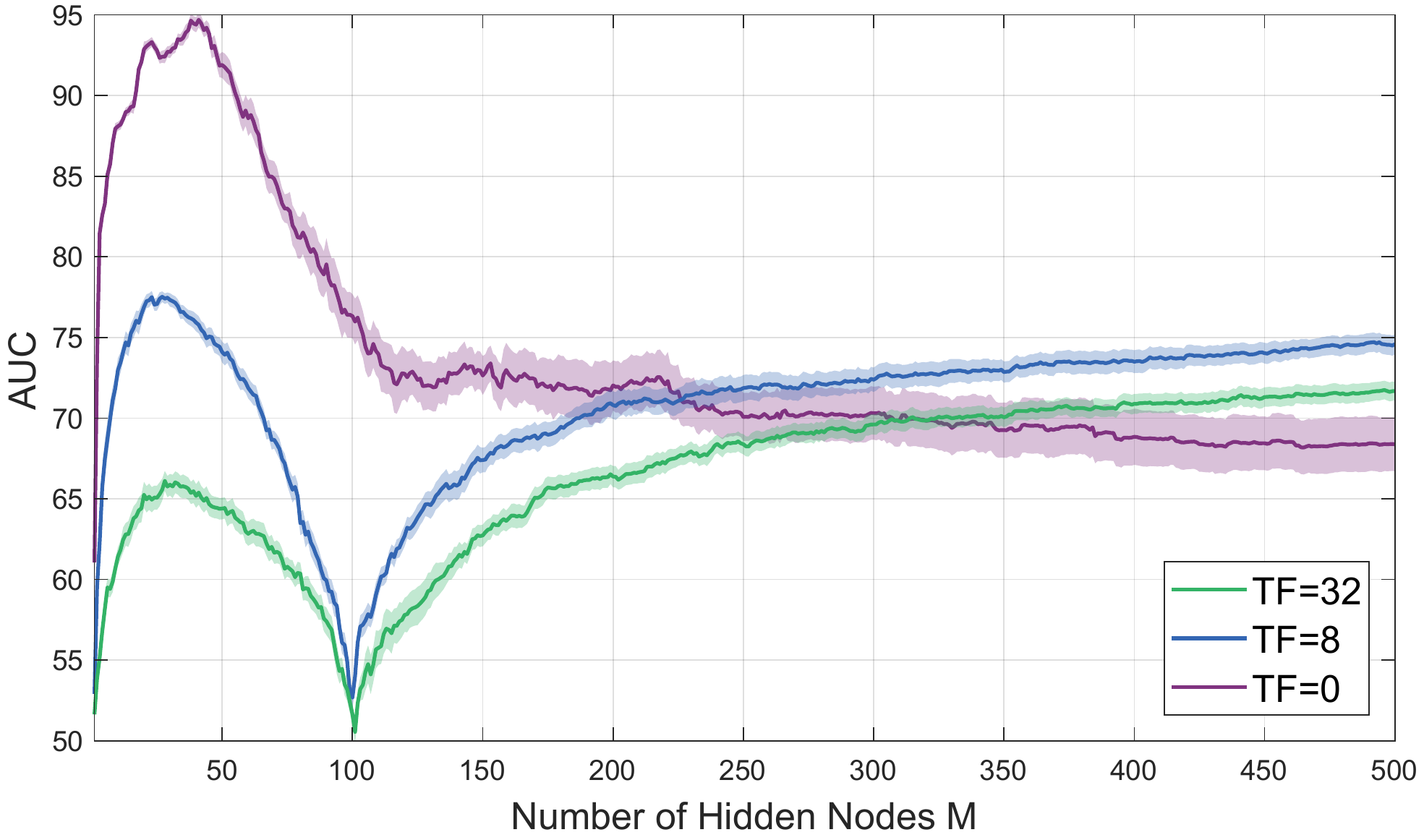}
\vspace{-0.5cm}
\caption{Mean test accuracy in synthetic data adding Junk Features (JF). Shading illustrates the standard deviation. }
\label{fig:sinhetic}       
\end{figure}

Let us take a look at Figure \ref{fig:sinhetic}, where for several choices of the number of neurons $M$, the average test accuracy of an ELM is plotted. The purple line corresponds to the test accuracy obtained using the data as displayed in Figure \ref{fig:lunitas}, while the others two correspond to those obtained using the datapoints contaminated with different numbers of junk features. One can immediately see that while a small choice of $M$ is optimal for the two-coordinates case, the more unrepresentative features the data contains, the more convenient it becomes to choose $M>>N$.

One might wonder, then, if this is just the result of the random choice of junk features, or if it is a scenario one might often expect in practical problems. To shed some light on this matter, we show an experiment using the DaSalla dataset (\cite{dasalla2009single}), which contains electroencephalography (EEG) signals recorded from three different subjects using 64 electrodes with a sampling frequency of 256 Hz. Here we used the EEG patterns related to imagination of mouth movement involved in the pronunciation of two vowels (/A/ and /U/). 

\begin{figure}
\includegraphics[width=\columnwidth]{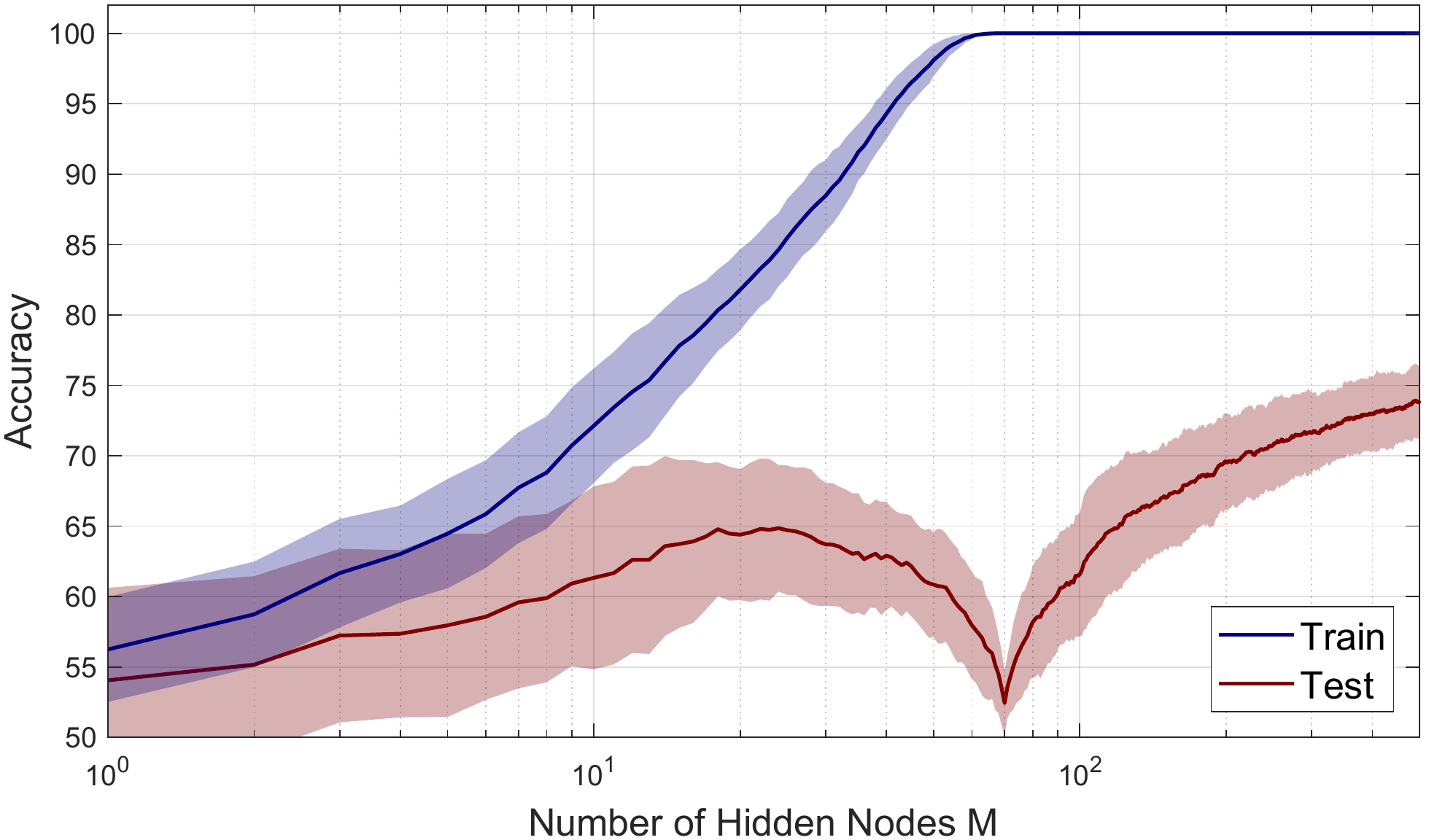}
\vspace{-0.5cm}
\caption{Mean accuracy obtained for training and testing data, as a function of the hidden layer size, $M$. Shading illustrates the standard deviation of the results.}
\label{fig:train_vs_test}       
\end{figure}

Figure \ref{fig:train_vs_test} illustrates the train and test accuracy of the ELM as a function of the parameter $M$. The experiments were performed using 70\% of the data for training and 30\% for testing, with 50 random initializations and 20 cross validations. As it can be seen, at first the accuracy grows with $M$, until reaching a local maximum, after which it starts to decay up to the global minimum, reached at $M=N$. However, the test accuracy starts to grow again as $M$ further increases. These results coincide with the findings in \cite{belkin2019reconciling}, where a similar behaviour is reported in the context of traditional multi-layer perceptrons.

\begin{figure}[ht]
\includegraphics[width=\columnwidth]{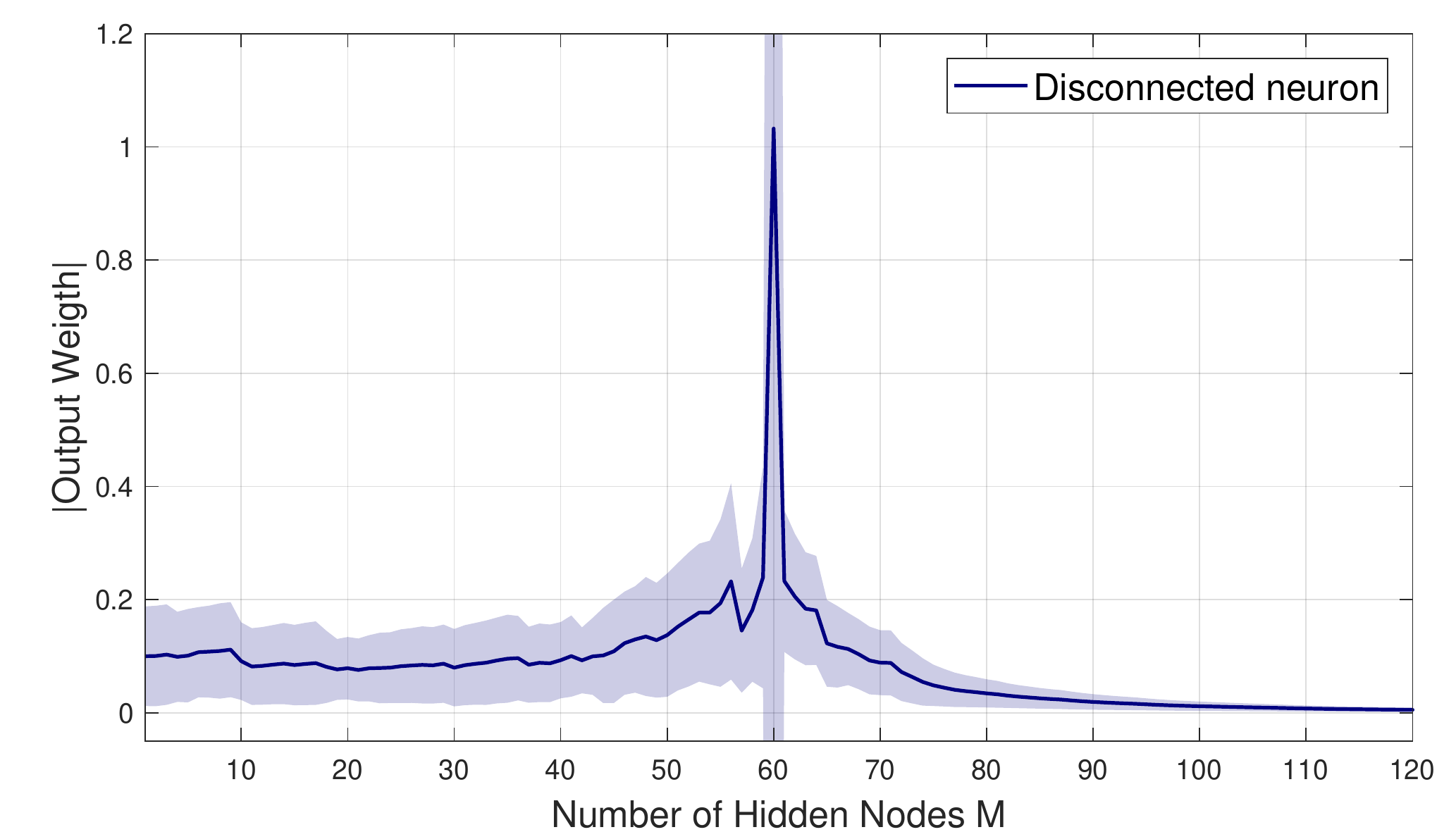}
\vspace{-0.5cm}
\caption{Mean absolute value of the weight of the neuron observing a fake feature as function of the hidden layer size (shading accounts for standard deviation over 50 initializations and 20 cross validations).}
\label{fig:false_beta}
\end{figure}

The overfitting observed in Figures \ref{fig:sinhetic} and \ref{fig:train_vs_test} when $M\approx N$ can be explained by the fact that when training under this condition, we are forcing the network to take into account all the features, even those irrelevant for classification. To corroborate that this is in fact the reason, we have performed an experiment consisting of adding a disconnected neuron to the ELM (before training using the DaSalla dataset). That is, a column of random elements having no correlation with the classes was stacked to the right of the matrix $H$. Figure \ref{fig:false_beta} depicts the absolute value of the weight that the model assigns to the disconnected neuron as a function of the number of neurons $M$. As seen, this weight remains small until $M$ approaches $N$, which supports our previous hypothesis. It is timely to observe, however, that the weight assigned to the fake neuron starts to decay again after this point. This means that for $M>N$, the ELM becomes capable of neglecting the value of the disconnected neuron, thus avoiding overfitting.

As we have shown, in certain cases, an ELM can benefit from choosing $M>N$. Yet this has the downside of increasing the network size. In order to find a compromise between the ELM size and its performance, one could use a validation method and retrain the network increasing the number of neurons $M$ until the change in performance is small enough to be neglected. Nevertheless, the computational burden associated to solving (\ref{eqn:b_hat}) increases with $M$, and the sensibility of the method with respect to the random initialization might require a few trials, making this idea unfeasible in practice. Hence, in order to make ELMs of practical use for real data as in the previous examples, in the next section, we introduce a pruning method that allows to reduce the network size after a single computation of (\ref{eqn:b_hat}) for $M>>N$. Comparisons between this pruning method, the standard $M<N$ setting, and a forward validation model in terms of performance and computational cost will also be shown.


\section{Relevance-based pruning}\label{sec:pruning}
Given that our proposal entails choosing $M$ significantly larger than $N$, it shall prove useful to have a method for reducing the ELM dimension. As it was observed in the description of Figure \ref{fig:false_beta}, the weight of a  neuron is proportional to the relevance of the corresponding feature. Hence, it is reasonable to ``throw away'' the neurons whose associated weights ($\beta$) are small enough. Given that the process of discarding a neuron and testing the performance of the resulting ELM is computationally inexpensive, the proposed pruning method begins using a large initial number of hidden nodes $M^*$, and then discard one (or a few) at a time until the performance exhibits a significant drop. We shall refer to the resulting method as Relevance-Based Pruning (RBP). The steps for doing this are shown in Algorithm \ref{alg:pruning}. In the next section  we show some experiments in order to compare RBP's performance against the traditional ELM approach.

\begin{algorithm}
\caption{: Relevance-Based Pruning (RBP)}
\label{alg:pruning}
\begin{algorithmic}
\STATE Set $M^*>>N$ and $\delta >0$.
\STATE Initialize the elements of $W \in \mathbb{R}^{M^*\times D}$ and $b\in \mathbb{R}^{M}$ as randomly with distribution $\mathcal{U}[-1,1]$.
\STATE $H = [g(W X_{train}+b\,1_{(1,N)})]\T$.
\STATE $\hat{\beta}= H^{\dagger}y_{train}$.
\STATE $H = [g(W X_{test}+b\,1_{(1,N)})]\T$.
\STATE Permute $\hat{\beta}$ so that $|\hat{\beta}_m|\geq|\hat{\beta}_{m+1}|, \;\forall m= 1, \ldots, M^*$. 
\STATE Perform the same permutation on the columns of $H$.
\STATE Let $H' = H$, $\hat{\beta}' = \hat{\beta}$
\WHILE{mean($H\hat{\beta}-y_{test})<\text{mean}(H'\hat{\beta}'-y_{test})+\delta$}
\STATE Let $H' = H$, $\hat{\beta}' = \hat{\beta}$
\STATE Remove the last element from $\hat{\beta}$.
\STATE Remove the last column from $H$.
\ENDWHILE
\end{algorithmic}
\end{algorithm}


\vspace{-0.2cm}
\section{Experiments}
\subsection{Experimental setting}
Two EEG datasets were used for the experiments. EEG data typically presents high levels of noise, and the relevant classification information is believed to be encoded in a particular subset of the features. The first is DaSalla dataset, already described in Section \ref{sec:netsize}. The second one is a P300-based BCI dataset \cite{base-datos}, consisting of 3780 EEG trials (630 with P300) acquired from 25 subjects using 10 channels at 256 Hz.


\subsection{Performance analysis}

In order to compare RBP against the standard ``growing'' ELM approach, we propose the following experimental setting: consider the datasets $X_{train}\in\mathbb{R}^{D\times N}$ and $X_{test}\in\mathbb{R}^{D\times N'}$ and a maximum ELM size $M^*$.
For the traditional method, the steps followed are shown in Algorithm \ref{alg:teststd}.

\begin{algorithm}
\caption{: Standard ELM growing scheme}
\label{alg:teststd}
\begin{algorithmic}
\STATE Set $M=1$.
\WHILE{$M<M^*$}
\STATE Initialize the elements of $W \in \mathbb{R}^{M^*\times D}$ and $b\in \mathbb{R}^{M}$ as randomly with distribution $\mathcal{U}[-1,1]$.
\STATE $H = [g(W X_{train}+b\,1_{(1,N)})]\T$.
\STATE $\hat{\beta}= H^{\dagger}y_{train}$
\STATE Compute the accuracy of the ELM using $X_{test}$ and $y_{test}$.
\STATE $M\leftarrow M+1$.
\ENDWHILE
\end{algorithmic}
\end{algorithm}

For illustrating the results of the RBP method, Algorithm was run \ref{alg:pruning} setting $\delta = 1$.

\begin{figure}
\includegraphics[width=\columnwidth]{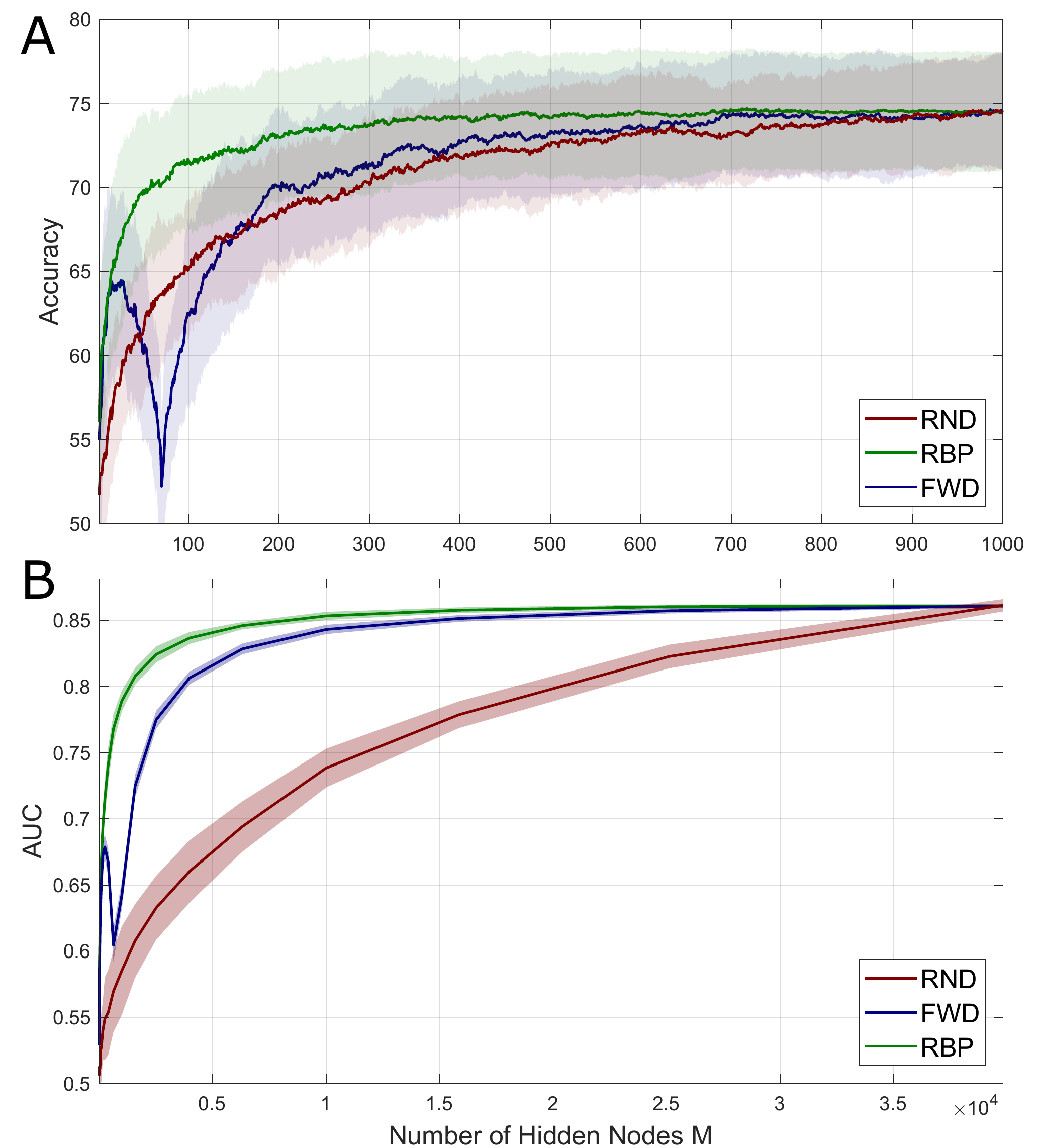}
\vspace{-0.5cm}
\caption{Average performance obtained with standard forward (FWD), Random Prunning (RND) and Relevance-Based Pruning (RBP) methods for one subject of (a) DaSalla dataset and (b) one subject of the P300 database. Shading accounts for standard deviation.}
\label{fig:dasalla_illustration}
\end{figure}

Using 50 random initializations over 20 cross validations for $W$ and $b$, for every value of $M\in\{1, \ldots, M^*\}$, we run the proposed experiment in a randomly chosen subject of the DaSalla dataset, with a 70/30 train/test scheme. The resulting average test accuracy for both the standard growing method (Algorithm \ref{alg:teststd}) and the RBP are depicted in Figure \ref{fig:dasalla_illustration}A. To validate the pruning criterion, the results obtained using a random pruning scheme (RND) are also shown.

An analogous test was made using a (randomly chosen) subject from the P300 database. Results are shown in Figure  \ref{fig:dasalla_illustration}B. Since, in this case the dataset and feature space are much larger, instead of taking increments of 1 on the values of $M$, a logarithmic grid was used. Also, given that the dataset is highly unbalanced, in order to evaluate performance we used the area under the ROC curve (AUC). The results correspond to five cross validations over 20 random initializations, with a choice of $M^* = 40000.$ 

As seen in Figure \ref{fig:dasalla_illustration}, RBP performs better than the forward scheme, and the average performance is (somewhat) monotonically increasing for $M>N$. This means that choosing a large $M^*$ followed by a reduction of the network size by means of RBP (until a significant decay in validation performance is observed) will yield a result at least as good as using the forward scheme. Thus, we can choose $M^*$ large enough (e.g. as determined by the computational cost we are willing to pay), and then run RBP to reduce the network size without loosing classification performance.

\subsection{Validation experiments}
In order to validate our proposal, we shall make comparisons between three different settings for each dataset:

\begin{description}[noitemsep]
    \item[STD:] (traditional approach) the best result obtained using $M<N$.
    \item[FWD:] start with $k=0$ and $M_k=1$. Increase $M_k$ over a logarithmic grid until the validation performance levels up or the maximum value $M^*$ is reached (see Algorithm \ref{alg:teststd}).
    \item[RBP:] start with $k=0$ and $M_k=M^*$ and decrease $M_k$ over a logarithmic grid until the first time that performance shows a significant (prescribed) reduction. The stopping criterion, grid and $M^*$ are set equal to those in FWD.
\end{description}

For our first experiment we take the full DaSalla dataset. We run twenty  cross validations, splitting the data in 70\%, 20\% and 10\% for training, validation and testing, respectively.  The parameters were set to $M^* = 1000$ and $\delta = 0.02$. Table \ref{tab:dasalla_results} shows the obtained results, from which two main issues can be observed: first, that both methods that allow for an ELM with a larger number of hidden nodes result in better performance, in terms of accuracy. Secondly, that between those two approaches, RBP is observed to required  much less computing time with a comparable network size.


\begin{table}
    \centering
    \begin{tabular}{|c|c c c|}
    \hline
    Method & Accuracy & time [ms] & $M$  \\
    \hline
    STD & $0.622 \pm 0.073$ & $\textcolor{white}{0}76  \pm 48$  & $\textcolor{white}{0}24  \pm  \textcolor{white}{0}18\,$  \\
    FWD & $0.669 \pm 0.055$ & $195  \pm 21$ & $228 \pm  \textcolor{white}{0}76$  \\
    RBP & $0.680 \pm 0.038$ & $\textcolor{white}{0}43  \pm \textcolor{white}{0}3$ & $209  \pm 151$  \\
    \hline
    \end{tabular}
    \vspace{-0.2cm}
    \caption{Experimental results using the DaSalla database.}
    \label{tab:dasalla_results}
\end{table}

\begin{figure}
\includegraphics[width=\columnwidth]{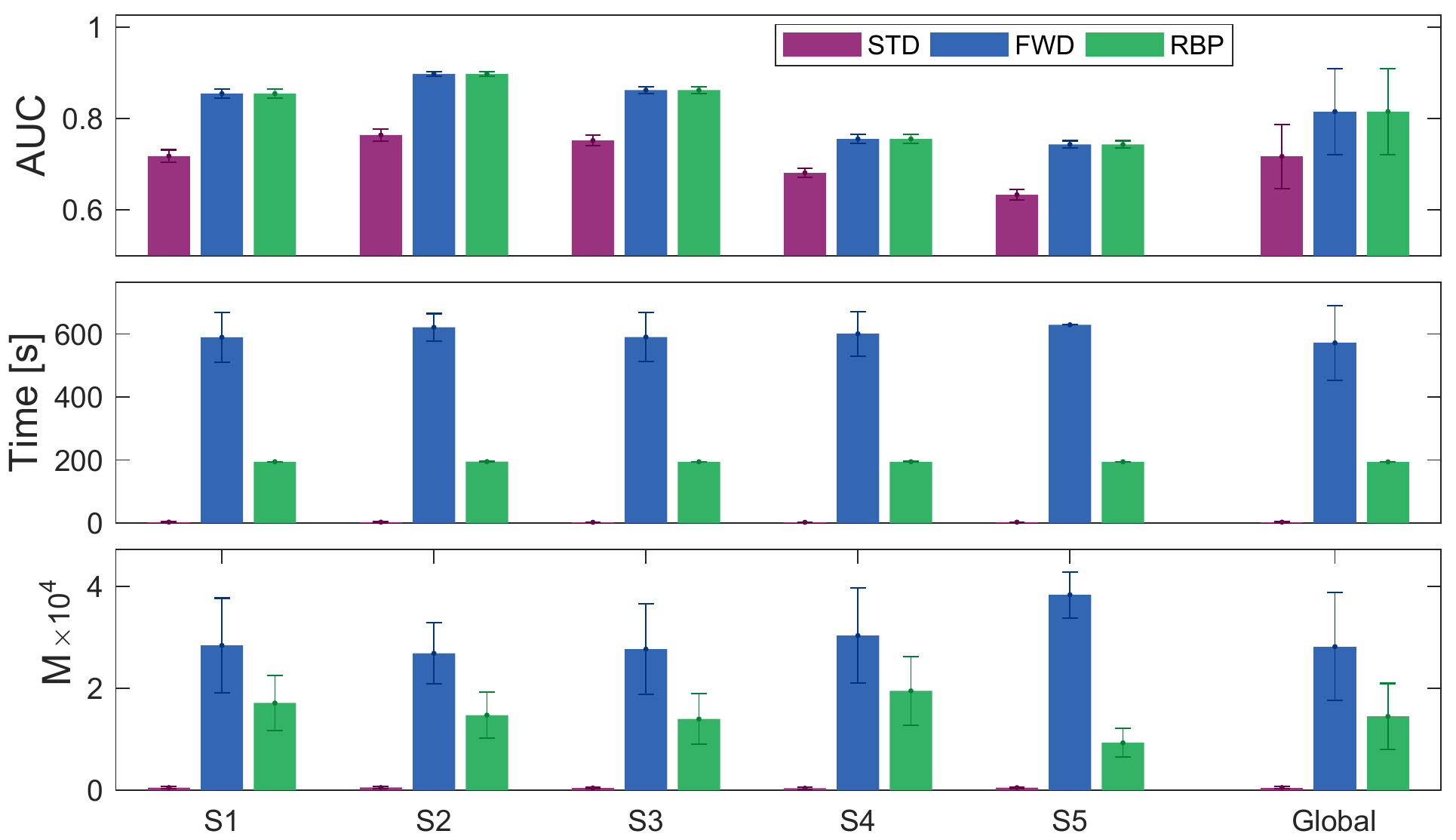}
\vspace{-0.5cm}
\caption{Results obtained with traditional (STD), forward (FWD) and Relevance-Based Pruning (RBP) methods on the P300 database.}
\label{fig:P300_results}
\end{figure}

The second experiment was performed using the P300-based BCI dataset. The data was split exactly as in the first experiment, for training, validation and testing, respectively. For every subject, we performed five cross validations over twenty random initializations. The parameters were set to $M^* = 40000$ and $\delta = 0.001$. Results are depicted in Table \ref{tab:p300_results}, and those obtained for the first five subjects are illustrated in Figure \ref{fig:P300_results}.

\begin{table}
    \centering
    \begin{tabular}{|c|c c c|}
    \hline
    Method & AUC & time [s] & $M _{{\times 10^3}}$  \\
    \hline
    STD & $0.718 \pm 0.072$ & $\textcolor{white}{00}3 \pm \textcolor{white}{00}1$  & $\textcolor{white}{0}0.5 \pm \textcolor{white}{0}0.2$  \\
    FWD & $0.816 \pm 0.096$ & $572 \pm 118$ & $28.2 \pm 10.6$  \\
    RBP & $0.816 \pm 0.087$ & $194 \pm \textcolor{white}{00}1$ & $14.5 \pm \textcolor{white}{0}6.5$  \\
    \hline
    \end{tabular}
    \vspace{-0.2cm}
    \caption{Experimental results using the P300 database.}
    \label{tab:p300_results}
\end{table}

As it can be seen, the AUC values obtained by FWD and RBP, i.e. the methods allowing for $M>N$, are significantly larger than those obtained with STD. While FWD and RBP do not account for a significant difference in AUC, RBP requires much less computing time than the former. Additionally, RBP yields a much smaller network size resulting in a more compact ELM.

From a practical point of view, if the number of hidden nodes is a restriction, then RBP is the best choice because it will yield better classification performance with the same number of neurons. On the other hand, if only classification is relevant, then RBP is more appropriate since it will yield the same performance in less or comparable training time than FWD with a more compact network.


\section{Conclusions and future work}
In this work we have shown that there are certain classification problems for which ELMs can benefit from using a non-traditional approach. Insights on why standard approaches are suboptimal (depending on the type of data) were provided along with a new pruning method to deal with these type of problems.

Results show that using a large number of hidden neurons can be beneficial and that the Relevance-Based Pruning method provides a time-efficient way to reduce the ELM size without jeopardizing performance. Furthermore, implementation is very simple and the benefits can be significant in some cases.

In the future, we shall tackle the problem of choosing an appropriate maximum ELM size ($M^*$) depending on the problem. Also, we intend to incorporate this method in the context of regularized ELMs and explore its potential as a feature selection tool.

\section*{Bibliography}

\bibliographystyle{spmpsci}      
\bibliography{bibliografia}   
%
%

\end{document}